# Globally Tuned Cascade Pose Regression via Back Propagation with Application in 2D Face Pose Estimation and Heart Segmentation in 3D CT Images

Peng Sun *pes2021@med.cornell.edu*
James K Min *jkm2001@med.cornell.edu*
Guanglei Xiong *gux2003@med.cornell.edu*
Dalio Institute of Cardiovascular Imaging, Weill Cornell Medical College

October 22, 2018

# Globally Tuned Cascade Pose Regression via Back Propagation with Application in 2D Face Pose Estimation and Heart Segmentation in 3D CT Images

**Peng Sun**
**James K Min**
**Guanglei Xiong**
Dalio Institute of Cardiovascular Imaging, Weill Cornell Medical College

PES2021@MED.CORNELL.EDU
EMAIL@COAUTHORDOMAIN.EDU
EMAIL@COAUTHORDOMAIN.EDU

## Abstract

Recently, a successful pose estimation algorithm, called Cascade Pose Regression (CPR), was proposed in the literature. Trained over Pose Index Feature, CPR is a regressor ensemble that is similar to Boosting. In this paper we show how CPR can be represented as a Neural Network. Specifically, we adopt a Graph Transformer Network (GTN) representation and accordingly train CPR with Back Propagation (BP) that permits globally tuning. In contrast, previous CPR literature only took a layer wise training without any post fine tuning. We empirically show that global training with BP outperforms layer-wise (pre-)training. Our CPR-GTN adopts a Multi Layer Percetron as the regressor, which utilized sparse connection to learn local image feature representation. We tested the proposed CPR-GTN on 2D face pose estimation problem as in previous CPR literature. Besides, we also investigated the possibility of extending CPR-GTN to 3D pose estimation by doing experiments using 3D Computed Tomography dataset for heart segmentation.

## 1. Introduction

Recently an effective technique for object pose estimation, referred to as *Cascade Pose Regression* (CPR), is proposed by computer vision community (Dollár et al., 2010) and has seen successful applications, e.g., the long line of work in face pose estimation (a.k.a. face alignment) (Cao et al., 2014; Burgos-Artizzu et al., 2013; Kazemi & Josephine, 2014; Ren et al., 2014; Xiong & De la Torre, 2013). Basically in CPR, the pose is represented as a set of *landmarks*. Then it learns a regressor that maps an unseen image to the coordinates of all the landmarks.

CPR is essentially a regressor ensemble, summing up a number of *stage regressors* which are learned in a stage-wise greedy-forward manner. Each stage regressor predicts a pose increment. CPR is thus very similar to Boosting, except that each stage regressor owns a private and unique *feature pool* that explicitly depends on the predicted pose up to its previous stage regressor. This way, CPR can gradually capture complicated pose and consequently be more robust to pose variations than a plain regressor.

In this paper, we investigate CPR from another perspective. We show how to formulate CPR as a *Neural Network* (NN) with sparse connections that "encode" our prior knowledge (i.e., the domain knowledge of image data). Specifically, we propose a *Graph Transformer Network* (GTN) representation (LeCun et al., 1998) so that CPR is trained globally with *Back Propagations* (BP). In contrast, the conventional CPR methods only adopt a layer wise training without any post fine tuning.

In previous work, the CPR is applied exclusively to 2D pose estimation. In this paper, we investigate the feasibility to solve 3D pose estimation problem by applying it to 3D *Computed Tomography* (CT) image for heart segmentation. See Fig.1.

### 1.1. Related Work

**Cascade Pose Regression**. Traditional image features, e.g., SIFT(Lowe, 2004), HoG (Dalal & Triggs, 2005), are blind to pose variation as they only depend on image pixel values. To alleviate this, (Fleuret & Geman, 2008) proposed the *Pose Index Feature* (PIF) that depends on both pixel values and the object pose. A simple yet effective PIF implementation is to model the pose with a set of landmarks and extract traditional image features in the neighbourhood of each landmark. However, the coupling of PIF and the pose estimation is a chicken-egg dilemma. (Dollár et al., 2010) thus devised an iterative algorithm alternating between pose estimation and PIF extraction with an initial pose guess, where the *random pixel difference* is adopted as PIF. In this line, much of the later work (Cao et al., 2014; Burgos-Artizzu et al., 2013; Ren et al., 2014; Kazemi & Josephine, 2014) showed that CPR can be successfully applied to an important computer vision problem – the face pose estimation (a.k.a. face alignment). Moreover, the *Supervised Descent Method* (SDM) (Xiong & De la Torre, 2013), motivated by non-convex optimization, can also be viewed as CPR where the PIF is based on HoG feature and the stage regressor is a linear SVM.

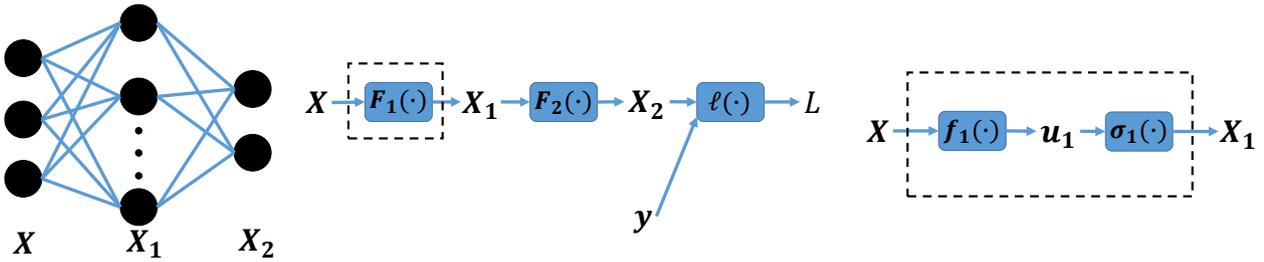

*Figure 2.* The MLP and its GTN representation. **Left**: conventional MLP graph. **Middle**: the equivalent GTN. **Right**: look into the GTN module surrounded by dashed rectangle, i.e., $F_1(\cdot)$.

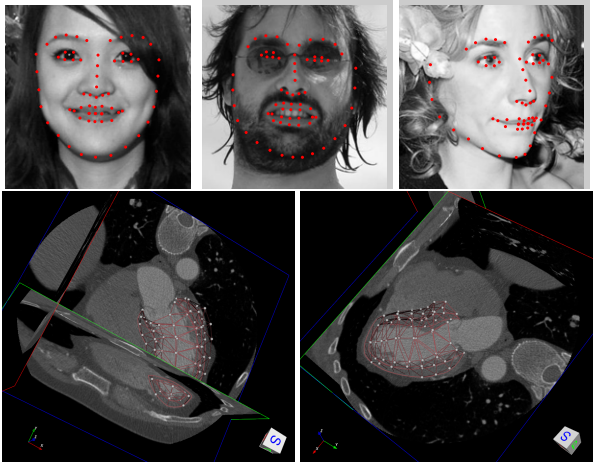

*Figure 1.* Examples of the datasets used in this paper. **Top**: face pose estimation from 2D image, landmarks in red; **Bottom**: heart (Left Ventricle) segmentation from 3D CT image, three mutually perpendicular 2D slices shown for each 3D image. Landmarks are shown in small white solid balls. The triangular connections among landmarks (red straight lines) are just for the purpose of display and are not used in our CPR training. Both the inner and outer surfaces are shown in meshes. Best be viewed on screen. See the texts in Section 4 for more explanations.

**CPR with NN**. In recent literature, NN was also proposed as the stage regressor for CPR. (Sun et al., 2013) and (Toshev & Szegedy, 2014) proposed to use *Convolutional Neural Network* (CNN) (Le et al., 2012) for face pose and human pose, respectively. (Zhang et al., 2014b;a) adopted Stacked Auto-Encoder.

**Layer-Wise (Pre-)Training**. All the aforementioned CPR methods, involving NN or not, train the regressor ensemble in a Boosting manner. In terms of NN training, the CPR is trained layer-wise, without any post fine tuning. In this work, however, CPR is trained globally with BP. As a result, an immediate question arises: can we take the layer wise training as pre-training which hopefully improves the performance of CPR? This issue is a little bit controversial in the NN literature. (Hinton & Salakhutdinov, 2006; Vincent et al., 2010) reported that unsupervised pre-training in a layer-wise way can improve the generalization of NN. (Girshick et al., 2014) showed that supervised pre-training on another datset with similar task can work well and unsupervised pre-training seems unnecessary. Moreover, (Ciresan et al., 2012) showed that no pre-training is needed at all provided the NN is carefully regularized and is appropriately deep enough. Given this ambiguous information conveyed in the literature, in this work we compared the three types of training (i.e., layer-wise training, layer-wise pre-training, pure BP) empirically and found that a pure BP worked best.

**Spatially Local Connection for NN**. In this work, we take the *Multi-Layer Perceptron* (MLP) as stage regressor, and we adopt a spatially local connection which is similar to the settings of CNN (LeCun et al., 1998) in order for local image feature representation. However, we don't use CN-N's weight sharing, which is inspired by the observations in recent literature (Le et al., 2012; Taigman et al., 2014). The underlying concept will be discussed in Section 3.5.

**Face Pose Estimation with Auxiliary Information**. When estimating face pose, auxiliary information can be helpful. Examples include: the rotation and scaling of head in 2D plane (Dantone et al., 2012; Smith et al., 2014) or 3D space (Cao et al., 2013), human identity (Chen et al., 2014; Cao et al., 2013), human gender and age (Zhang et al., 2014c), etc. Subsequently, the pose regressor training is formulated as multi-task learning or similar framework. These methods are shown to significantly improve the accuracy of pose estimation. However, we will not consider them. In this paper we'll keep our focus on the connection between CPR and NN.

**CPR for 3D CT Image**. Previous CPR work only studied the pose estimation for 2D images. In this work, we investigate the feasibility of applying CPR to 3D pose estimation for 3D CT image segmentation, where the previous work in medical image processing literature usually adopted an ASM/AAM based method (Zheng et al., 2008) or segmentation/edge-detection based method (Ecabert et al., 2008).

### 1.2. Summary and Outline

The technical contributions of this paper can be summarized as follows:

• We show how a CPR can be represented by a GTN. The derivatives of each GTN module, particularly the PIF extraction module, is provided.

• We show CPR can be trained with pure BP algorithm, which outperforms layer-wise training (the way previous work on CPR adopts) or layer-wise pre-training in our experiments.

• We show CPR stage regressor can be a *Multi-Layer Perceptron* (MLP), which can benefit from spatially local connection that learns local image feature representation.

• We show the CPR-GTN can be extended to the problem of 3D pose estimation from 3D CT image.

The rest of this paper is organized as follows: in Section 2 we briefly review GTN. In Section 3 we first review CPR and then show how it is formulated as GTN. In Section 4 we show experiments on a public face pose dataset and on a 3D CT image dataset for heart segmentation.

## 2. Review of Graph Transformer Network

The concept of Graph Transformer Network (GTN) is well described in Section IV of (LeCun et al., 1998). For completeness of this paper, we briefly review it in this section with slight modifications on notations.

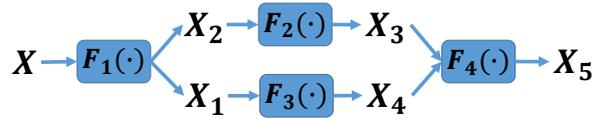

*Figure 3.* A general GTN.

Basically, GTN is a general representation of deep model, which includes, but not confines to, feed-forward neural network with bi-directional list structure. Think of a simple MLP with one hidden layer as an example. The traditional graph is usually drawn as in the left of Fig.2. However, it can also be given in GTN form as in the middle of Fig.2. The module $F_1(\cdot)$ is a wrapper, which itself can be expanded to sub modules $f_1(\cdot)$ and $\sigma_1(\cdot)$ as in the right of Fig.2 (Module $F_2(\cdot)$ goes similarly.).

Each module transforms its input variable to the output one. In this example, $f_1(\cdot) : \boldsymbol{x} \mapsto \boldsymbol{u}_1$ is a linear transformation, while $\sigma_1(\cdot) : \boldsymbol{u}_1 \mapsto \boldsymbol{x}_1$ performs a point-wise activation (e.g., sigmoid or tanh). $F_1(\cdot)$ is the composition of $f_1(\cdot)$ and $\sigma_1(\cdot)$ ($F_2(\cdot)$ goes similarly). Finally, $\ell(\cdot) : (\boldsymbol{x}_2, \boldsymbol{y}) \mapsto L$ calculates the loss where $\boldsymbol{y}$ is the label associated with the instance and the loss $L \in \mathbb{R}^+$ is a non negative scalar. Note that the module $\ell(\cdot)$ is only used in training while it's unnecessary in testing. To this extent, the feed-forward or back-propagation procedure for GTN can be seen as the message passing over the variables $\boldsymbol{x}, \boldsymbol{u}_1, \boldsymbol{x}_2..., \ell$ in either forward or backward order. Therefore, the training and testing of GTN go exactly the same with the "traditional" graph drawn in the left of Fig.2.

In general case, however, GTN can be an Directed Acyclic Graph (DAG), permitting the representation of more complicated deep model. What's interesting is that feed-forward or back-propagation procedure still goes well with DAG if we restrict the variable order to be consistent with the parent-child relationship. Consider Fig.3 as an example. When feed-forwarding, the transformation $F_1(\cdot)$ must be prior to $F_2(\cdot)$, while the transformation $F_2(\cdot)$ and $F_3(\cdot)$ can be in parallel. Similar requirement applies to back propagation.

## 3. The Proposed Method

In this section we show how to represent CPR by GTN so that CPR can be trained with Back Propagation. We begin with a brief review of CPR.

### 3.1. Review of Cascade Pose Regression

We demonstrate how CPR goes with the example of face pose estimation. A particular face pose is modeled as a set of $L \in \mathbb{N}^+$ landmark points locating at eye corner, mouth corner, chin, etc. Denote the pose by $\boldsymbol{p} \in \mathbb{R}^{2L}$ which is a vector concatenating the $x$-$y$ coordinates of the $L$ points. Typically, $L = 68$. For a given image $I \in \mathbb{R}^{W \times H}$ with the face fitting the image size, we train a regressor $R(\cdot)$ : $\mathbb{R}^{W \times H} \mapsto \mathbb{R}^{2L}$ to predict the face pose: $\boldsymbol{p} = R(I)$.

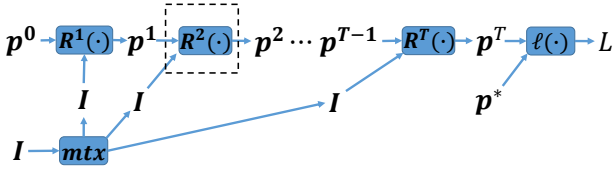

*Figure 4.* The CPR as a GTN

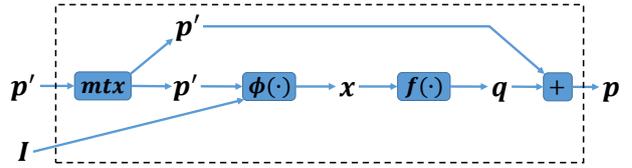

*Figure 5.* The stage regressor $R^t(\cdot)$.

Given the complicated pose variations including face expressions, geometrical transformation, lighting, etc., learning the regressor $R(\cdot)$ directly is not easy. (Dollár et al., 2010) proposes to accomplish this in a cascaded manner so that the pose is predicted in a gradually refined way. Specifically, the method, called CPR, starts with an initial pose guess $\boldsymbol{p}^0$. Inevitably, there is a difference, called residual hereinafter, $\Delta \boldsymbol{p} = \boldsymbol{p}^* - \boldsymbol{p}^0$ between $\boldsymbol{p}^0$ and the ground truth pose $\boldsymbol{p}^*$. Then, a regressor $R^1(I, \boldsymbol{p}^0)$, taking as input both image $I$ and a pose $\boldsymbol{p}^0$, is fitted with least square to predict the residual $\Delta \boldsymbol{p}$. After this is done, hopefully the guess is updated to be a more accurate prediction $\boldsymbol{p}^1 \leftarrow \boldsymbol{p}^0 + R(I; \boldsymbol{p}^0)$ while the residual is reduced to be $\Delta \boldsymbol{p} \leftarrow \boldsymbol{p}^* - \boldsymbol{p}^1$. Subsequently, based on $\boldsymbol{p}^1$ we train $R^2(I; \boldsymbol{p}^2)$, and so on. We end up with a combination of $T \in \mathbb{N}^+$ regressors:

$$R(I) = \boldsymbol{p}^0 + \sum_{t=1}^{T} R^t(I, \boldsymbol{p}^{t-1}), \qquad (1)$$

where the pose is updated incrementally

$$\boldsymbol{p}^t = \boldsymbol{p}^{t-1} + R^t(I, \boldsymbol{p}^{t-1}) \qquad (2)$$

for *stage* $t = 1, 2, ..., T$. In equation (1) or (2), $R^t(I, \boldsymbol{p}^{t-1})$ is referred to as *stage regressor*, which is learned at stage $t$ based on the pose estimation $\boldsymbol{p}^{t-1}$ up to last stage.

### 3.1.1. Pose Index Feature

The stage regressor $R^t(I, \boldsymbol{p}^{t-1})$ works over the so called Pose Index Feature (PIF). Differen from conventional image feature depending only on image $I$, the PIF depends on both image $I$ and the pose estimation $\boldsymbol{p}^{t-1}$ at last stage. A simple yet effective implementation is to first extract the conventional features in a small neighborhood of each pose landmark and then concatenate/pool them. This way, the feature would be hopefully invariant to pose variations and thus be more reliable. (Cao et al., 2014; Ren et al., 2014; Burgos-Artizzu et al., 2013; Kazemi & Josephine, 2014) utilized Random Pixel Difference as underlying conventional features, while (Xiong & De la Torre, 2013) resorted to

HoG. In this work we simply adopt the Random Pixel Difference feature, which means taking many random pointpairs and concatenating the difference values.

### 3.1.2. Layer-Wise Learning

(Dollár et al., 2010) proposed to learn CPR in a forward, greedy stage wise way that is similar to Boosting (Friedman et al., 2000). For a regressor consisting of $T$ stage-regressors, it iterates exactly $T$ times and learns just one stage regressor per iteration based on those regressors learned in previous iterations. Specifically, at iteration $1 \le t \le T$, only the stage regressor $R^t(I; \boldsymbol{p}^{t-1})$ is learned over the features extracted based on $\boldsymbol{p}^{t-1} = \boldsymbol{p}^0 + \sum_{t'=1}^{t-1} R^{t'}(I, \boldsymbol{p}^{t'-1})$, which implicitly depends on all previously learned regressors. This is almost Boostingstyle learning except that the feature pool is changed from one iteration to another due to the introduction of PIF. This type of learning for CPR is followed by other authors (Cao et al., 2014; Ren et al., 2014; Burgos-Artizzu et al., 2013; Kazemi & Josephine, 2014). The SDM (Xiong & De la Torre, 2013), although motivated by pure optimization issue other than CPR, also learns in such a layer-wise manner.

### 3.2. CPR as GTN

In this subsection we show how the CPR in last subsection can be formulated as a GTN. Recalling equation (1), (2) and using the graphic notations reviewed in Section 3.1, we can express the regressor (1) by a GTN shown in Fig.4.

We now explain each variable, each module and associated transformation in Fig.4. The loss module $\ell(\cdot)$ calculates the least square error between the pose estimation $\boldsymbol{p}^T$ at last stage and the ground truth pose $\boldsymbol{p}^*$. The module "mtx" is a multiplexer that replicates its input at each of its output. The module $R^t(\cdot)$ is the stage regressor (1), taking as input both image $I$ and pose estimation $\boldsymbol{p}^{t-1}$ and outputting the pose estimation $\boldsymbol{p}^t$. Looking inside, the $R^t(\cdot)$ can decompose into the sub-modules as in Fig.5.

In a slight abuse of notations and without any confusion, we have renamed the input and output variables in Fig.5 for simplicity and avoiding name conflict. The "mtx" is again a multiplexer. The module "+" takes the sum of the input pose and the predicted residual: $\boldsymbol{p}' = \boldsymbol{p} + \boldsymbol{q}$. The module $\phi(\cdot)$ is a PIF extractor that takes as input the current pose estimation $\boldsymbol{p}$ as well as the raw image $I$. The output feature $\boldsymbol{x}$ is fed into a regressor $f(\cdot)$ to predict the pose residual $\boldsymbol{q}$. In the following we give more details on $\phi(\cdot)$ and $f(\cdot)$.

As is described in Section 3.1.1, the PIF $\boldsymbol{x} = \phi(I, \boldsymbol{p})$ can be simply built based on a conventional image feature by restricting the feature extraction in the neighborhood of each landmark and then concatenating them. In this study we follow the line of (Cao et al., 2014; Ren et al., 2014; Burgos-Artizzu et al., 2013; Kazemi & Josephine, 2014) and adopt the random pixel difference features. As for the regressor $f(\cdot)$, we let it be an MLP with only one hidden layer. However, we only allow spatially local connection for the hidden layer instead of the full connection as in conventional MLP, which will be discussed in Section 3.5 in greater details.

### 3.3. Training with Back Propagation

As is pointed out in (LeCun et al., 1998), the GTN can be well trained with the classical BP algorithm if each module is differentiable w.r.t. its input variables and parameters (if any) almost everywhere. This is the case for the CPR-GTN proposed in this paper, as is explained in the following.

Since the repeated structure of CPR-GTN as in Fig.4, let's just check how BP goes for $R^t(\cdot)$ as in Fig.5. $f(\cdot)$ is an MLP, whose BP is standard in textbook, e.g., (Bishop et al., 1995), even in the case of sparse connection (See Section 3.5). The BP for module "mtx" is at input taking the sum of the delta-signals at each output, while the BP for module "+" is at each input taking the replication of the delta-signal at the output. This symmetrical criterions are described in (LeCun et al., 1998). The only thing that needs a little bit more explanations is the BP for the PIF module $\phi(\cdot)$, which is given in the followed contents.

Keep Fig.4 in mind, let's start from the simplest case. Let there be only one landmark such that $\boldsymbol{p} \in \mathbb{R}^2$, each entry corresponding to horizontal and vertical coordinate respectively. Let the randomly picked two points deviate from $\boldsymbol{p}$ by $d_1 \in \mathbb{R}^2$ and $d_2 \in \mathbb{R}^2$, respectively. In a slight abuse of notation and without confusion, for image $I$ we let the scalar random pixel difference feature $x$ be

$$x = \phi(\boldsymbol{p}, I) = I(\boldsymbol{p} + d_1) - I(\boldsymbol{p} + d_2), \tag{3}$$

where $I(\boldsymbol{p})$ denotes the image pixel value at point $\boldsymbol{p}$. From (3) we have the derivatives w.r.t. $\boldsymbol{p}$ as:

$$\frac{\partial x}{\partial \boldsymbol{p}} = \nabla I(\boldsymbol{p} + d_1) - \nabla I(\boldsymbol{p} + d_2), \tag{4}$$

where $\nabla I(\boldsymbol{p}) \in \mathbb{R}^2$ denotes the 2D image gradient vector at point $\boldsymbol{p}$. Likewise, we have the derivatives w.r.t. $I$ as:

$$\frac{\partial x}{\partial I} = \delta(\boldsymbol{p} + d_1) - \delta(\boldsymbol{p} + d_2), \tag{5}$$

where $\delta(\boldsymbol{p})$ denotes the pulse response at point $\boldsymbol{p}$, i.e., it takes value 1 at $\boldsymbol{p}$ and value 0 at other position. Note that (5) is a matrix with the size of image $I$.

Now, in Fig.4 denote by $L$ the loss and denote by $\frac{\partial L}{\partial x}$ the scalar delta-signal propagated into the output. According to chain rule in calculus, we have the delta-signal propagated to input variables $\boldsymbol{p}$ and $I$ to be

$$\frac{\partial L}{\partial \boldsymbol{p}} = \frac{\partial x}{\partial \boldsymbol{p}} \frac{\partial L}{\partial x} \tag{6}$$

and

$$\frac{\partial L}{\partial I} = \frac{\partial x}{\partial I} \frac{\partial L}{\partial x}, \tag{7}$$

respectively, where we should substitute (4) and (5) for concrete calculations. Note that $\frac{\partial L}{\partial \boldsymbol{p}} \in \mathbb{R}^2$ and $\frac{\partial L}{\partial I}$ is with the same size of image $I$.

In general case, there are $L > 1$ landmarks. Moreover, in the neighborhood of each landmark there can be $M > 1$ random point pairs. Therefore, equation (4), (5), (6), (7) should take appropriate matrix form. The mathematical details go similarly and are omitted here.

**Remark 1.** The smoothness of gradient. For a discrete digital image, the pixel value $I(\boldsymbol{p})$ is not strictly differentiable w.r.t. to the position $\boldsymbol{p}$. Therefore, $\nabla I(\cdot)$ in equation (4) can only be calculated approximately. A common technique in image processing is the 2D convolution over image $I$ with a carefully designed template that simultaneously mimics the gradient and enforces smoothness. In this paper we simply adopt the Sobel template (Gonzalez et al., 2004).

**Remark 2.** $\phi(\boldsymbol{p}, I)$ being vector features. In the case of one landmark, the random pixel difference feature $x$ in (3) is the subtraction of raw image pixel values and is thus a scalar. However, the feature can also be a vector $x \in \mathbb{R}^D$ with $D \in \mathbb{N}^+$. To calculate equation (4) in the case of vector, we can simply view $x$ as an image with $D$ channels and calculate the 2D gradient independently for each of the channel. For instance, suppose we are building $x = \phi(I, \boldsymbol{p}) \in \mathbb{R}^{128}$ on top of HoG-like features (Dalal & Triggs, 2005; Xiong & De la Torre, 2013) which extracts the gradient histogram for 8 orientations over each of the $4 \times 4$ grids ($D = 128 = 8 \times 4 \times 4$) on a rectangle image patch centered at some landmark $\boldsymbol{p}$. To obtain $\frac{\partial x}{\partial \boldsymbol{p}} \in \mathbb{R}^{2 \times 128}$, we can first calculate dense HoG over every point $\boldsymbol{p}$ in image and then calculate the 2D Sobel gradient over each of the 128 channels independently. We end up by fetching the results at the interested point $\boldsymbol{p}$ and concatenating the 128 2D gradient vectors.

**Remark 3.** The derivative w.r.t. $I$ for visualization. At first glance, the derivative (5) and (7) seem useless. For a certain training image $I$, it does not make sense to change its pixel values when training so that the delta-signal for $I$ can be ignored. However, (5) and (7) are indeed useful for other purpose, e.g., the visualization which will be discussed later in this section.

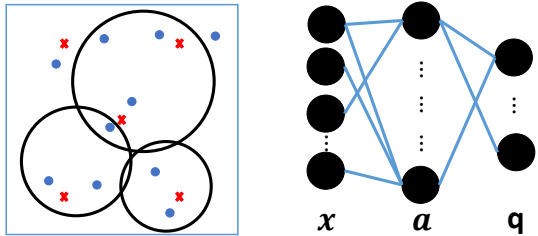

*Figure 6.* Spatially local connection. **Left**: landmarks (red crosses), features (blue points) and locality (circle). **Right**: the corresponding MLP. See texts for explanations.

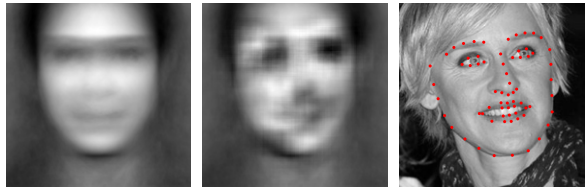

*Figure 7.* The synthesized image for a ground truth pose. **Left**: the average image as initialization; **Middle**: the most likely image; **Right**: the ground truth pose with its face image.

Finally, in Fig.4 the feed forward procedure for all the modules are straightforward. To this extent, we've got all the details to train CPR-GTN with BP. The parameters can be updated using *Stochastic Gradient Descent* (SGD), as is discussed in Section 3.6.

### 3.4. Pure BP and Layer-Wise (Pre-)Training

In this sub section we discuss more on CPR training. In last subsection we've shown it can be trained with pure BP. In the meanwhile, in previous computer vision literature (Dollár et al., 2010; Burgos-Artizzu et al., 2013; Cao et al., 2014; Kazemi & Josephine, 2014) the CPR is learned in a layer-wise way that one layer is added once a time.

On the other hand, in recent NN literature it is believed that unsupervised (Hinton & Salakhutdinov, 2006; Vincent et al., 2010) or supervised (Girshick et al., 2014) pre-training is a critical technique that improves the performance of deep NN structure. For example, a Restricted Boltzman Machine and Stacked Auto Encoder can be learned in a layer-wise manner. Then the structure is fixed and the parameters are preserved, serving as the initialization of a further "fine-tuning" by using BP. This strategy is believed to be better than a pure BP with random parameter initialization. However, there is also later work (Ciresan et al., 2012) showing that pre-training is not needed at all provided the deep structure is regularized carefully and trained sufficiently. It is still unclear whether a pre-training should be taken or not – a convincing theoretical explanation seems absent in the literature.

Given the short discussion above, for the CPR-GTN in this paper we are immediately facing three types of training: 1. A pure BP with random initialization; 2. A layer-wise training as in aforementioned computer vision literature; 3. A layer-wise pre-training that is in-between 1 and 2, i.e., first perform layer-wise training, then take it as initialization and continue training with BP. Accordingly, we compare the three types of training empirically and find that the pure BP performs best, see Section 4.1.

### 3.5. Spatially Local Connection

In Fig.5, the module $f(\cdot) : \boldsymbol{x} \mapsto \boldsymbol{q}$ can be arbitrary regressor provided it is differentiable w.r.t. its input and parameters which permit BP [1]. An immediate choice seems a fully connected MLP. In this paper we adopt a single hidden layer MLP such that $\boldsymbol{a} = \sigma(\boldsymbol{W}_1 \boldsymbol{x} + \boldsymbol{b}_1)$ and $\boldsymbol{q} = \boldsymbol{W}_2 \boldsymbol{a} + \boldsymbol{b}_2$, where the hidden variable $\boldsymbol{a}$ can be viewed as an intermediate feature representation that has higher abstraction than feature $\boldsymbol{x}$. However, it is well known that fully connected MLP tends to be over-fitting due to its too large capability of representation. To alleviate it, we could restrict the connection between $\boldsymbol{x}$ and $\boldsymbol{a}$ to be spatially local.

We explain the spatially local connection by an example as in the left of Fig.6. Suppose there are $L$ landmarks (red crosses) and $M$ features (blue points), each indicating the middle point of a random point-pair) such that we have in total $\boldsymbol{x} \in \mathbb{R}^{N_1}$ where $N_1 = ML$. Now look at the hidden layer $\boldsymbol{a} \in \mathbb{R}^{N_2}$ for some $N_2 \in \mathbb{N}^+$. For the first component of $\boldsymbol{a}$, we denote it by the centroid of a circle and let it only connect to those blue points inside the circle. The other $N_2 - 1$ components of $\boldsymbol{a}$ go similarly. The up to $N_2$ circles should scatter reasonably and their union should approximately cover all the $N_1$ blue points. The corresponding MLP is drawn as in the right of Fig.6, where we resort to conventional graphic notations of feed forward network to emphasize its sparse connection intending for local feature representation. In this paper we simply let $\boldsymbol{q}$ to be fully connected to $\boldsymbol{a}$, although local connection seems also make sense.

**Remark.** Our choice for $\boldsymbol{a}$ is similar to the "feature map" produced by a convolutional layer in CNN. However, there is noticeably difference:

- First, the convolution in CNN directly applies to every raw image pixels, while our $\boldsymbol{a}$ takes as input the random pixel difference features – such high level features are spatially scattered by construction so that they are very likely to be mutually uncorrelated and to be complementary. In contrast, CNN needs additional sub-sampling (or pooling) layer to eliminate the correlations for the feature map, or it would compromise to noises of close pixel and hence-forth lose the property of translation-invariant feature representation.

---

[1] Therefore, we have to exclude Tree regressor which is popular in Boosting or Random Forest but is not differentiable.

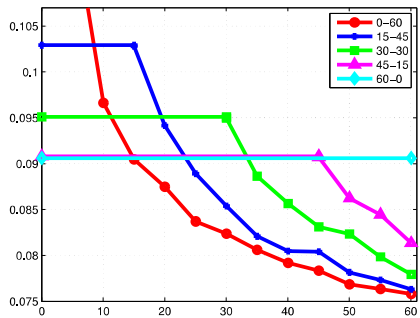

Figure 8. Various types of training: epoches vs testing error.

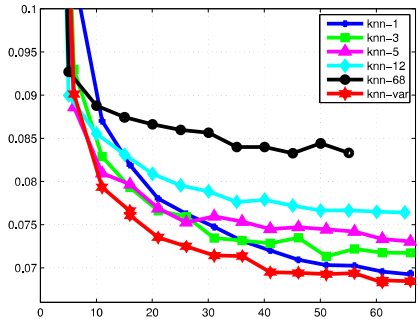

Figure 9. Various local connections: epoches vs testing error.

• Second, the convolution in CNN actually adopts both spatially local connection and weight sharing. In this paper, however, the weight sharing is explicitly dismissed when constructing $\boldsymbol{a}$. This technique was proposed in previous literature (Taigman et al., 2014; Le et al., 2012) and the considerations are explained therein. We rephrase them here. Suppose the task of face recognition(Taigman et al., 2014). The features having high response for eyes should be intuitively different with that for mouth, therefore it suffices to let the filter kernel move in a small image patch and thus yield a corresponding small feature map (or even let it be fixed on a single pixel, yielding feature map with only one element (Le et al., 2012), which is exactly our choice in this paper). This would enforce the specificity of the feature representation, i.e., it helps produce features that are specialized in pupil, eyebrow, tooth, etc.

### 3.6. Implementation Details of Training

We discuss the geometrical transform of pose, the data augmentation trick and the parameter tuning for SGD. See the supplement for details.

### 3.7. Visualization: Synthesize Image from Pose

As a bonus of GTN representation, the trained CPR is rather easy to visualize (Simonyan et al., 2013), that is, for a given ground truth pose $\boldsymbol{p}^*$, what's the most likely image $I$ the regressor $R(\cdot)$ is fed so that $\boldsymbol{p}^* = R(I)$? An example of the synthesized image for a given pose is shown in Fig.7. Details are put in supplement.

## 4. Experiments

We did experiments on a public 2D face pose dataset called 300-W and a 3D pose estimation dataset in CT images for hear segmentation, as described in the following.

**300-W (testing on "common subset")**. The source is plain 2D images of human face, where the face pose is modeled by a set of $L = 68$ landmarks (Sagonas et al., 2013). The 300-W dataset consists of several subsets (LFPW, Helen, AFW, etc.) which are publicly available at http://ibug.doc.ic.ac.uk/resources/facial-point-annotations/. In this paper, we used the full training set containing 3148 images, while we tested on the so-called "common subset" defined in (Ren et al., 2014) including 689 images. Examples are shown in Fig.1.

**CT Images**. The source is 3D CT images, each image in the size $X \times Y \times H$ where $X = Y = 512$ and $H$ ranges from 150 to 210. The $X$-$Y$ plane resolution ranges from 0.33 to 0.43 millimeters per pixel while the $Z$ axis resolution ranges from 0.60 to 0.62 millimeters per pixel. We are interested in segmenting the left ventricle (LV) of the heart[2]. The LV is a bowl-like cavity, therefore we need delineate both an LV's inner surface and outer surface, as is in Fig.1. We approach this by pose estimation based on supervised learning. Specifically, we adopt $L = 172$ landmarks (86 for the inner surface and 86 for the outer surface) to fully capture an LV's pose in a CT image. We randomly assign 112 CT images for training and 27 for testing, with each's landmarks labeled by a human-expert.

### 4.1. Pure BP vs layer-wise (Pre-)Training

In thie subsection we compare three types of training: BP, layer-wise training and layer-wise pre-training, whose mutual relation has been discussed in Section 3.4. We intend to answer such a question: which type of training is most efficient? Or equivalently, how fast does each type of train-ing decrease the testing error?

---

[2]A heart has four chambers, LV is the most interesting chamber to the clinical doctors.

*Table 1.* Testing error on 300-W.

| LBF | LBF fast | SDM | This Paper |
|------|----------|------|------------|
| 4.95 | 5.38 | 5.60 | 6.76 |

*Table 2.* Mesh distance (in millimeters) for heart (Left Ventricle) segmentation.

| | AAM/ASM | This Paper | Do Nothing |
|---------------|---------|------------|------------|
| inner surface | 1.13 | 3.35 | 7.24 |
| outer surface | 1.21 | 3.41 | 7.31 |

We set a CPR with $T = 8$ stages (layers). We run the three types of training on 300-W dataset and fixed the total number of SGD iterations to be 60 epoches. We distributed these 60 epoches to layer-wise pre-training and the following BP in different ways to figure out its effect on the testing error decrease. Results are reported in Fig.8, in whose legend the format "*m-n*" means $m$ epoches of layer-wise pre-training followed by $n$ epoches of BP. Therefore, 60-0 means only the layer-wise training with each layer trained by 60 SGD epoches; 0-60 means only the 60 epoches of BP with random initialization; 15-45, 30-30 and 45-15 are all in-between layer-wise pre-training entered. Note that in Fig.8 the layer-wise training is simply plotted as flat curve, which should not lead to any confusion.

The testing error reported in Fig.8 is the so called *pupil-distance normalized mean distance* (in percentage %), which is a widely adopted measure in the face pose estimation literature (Ren et al., 2014) (Xiong & De la Torre, 2013) (Dantone et al., 2012) (Burgos-Artizzu et al., 2013). The same measure is reported for all our experiments on face pose.

As we can see, BP's curve is almost below all the others, which suggesting that BP decreases the fastest. The testing error decreases slower and slower with more and more epoches of layer-wise pre-training entered. Finally, an absolute Layer-wise training performs the worst.

### 4.2. The Spatially Local Connection

As is discussed in Section 3.5, the spatially local connection is beneficial. In this subsection, we investigate the impact by varying the locality of connections as follows.

We let the regressor $f(\cdot)$ be a one hidden layer MLP. For a pose with $L$ landmarks, we extracted $M$ features around each landmark, constituting $L \times M$ features: $\boldsymbol{x} \in \mathbb{R}^{LM}$. For the intermediate feature map representation $\boldsymbol{a}$ (i.e., the hidden layer of MLP), we let its number of features (i.e., the number of neurons) be $\boldsymbol{a} \in \mathbb{R}^{LH}$ with $H < M$ to learn a compact feature representation. By this settings, the components of $\boldsymbol{a}$ can be exactly divided into $L$ groups, each group corresponding to one landmark. For a feature in $\boldsymbol{a}$ associated with landmark $i$, we let it connect to those $\boldsymbol{x}$ features who are from landmark $i$'s $k$-nearest-neighbor landmarks, i.e., each feature in $\boldsymbol{a}$ connect to $Lk$ features in $\boldsymbol{x}$. In our experiments, $k \in \{1, 3, 5, 12, 68\}$, as in the legend of Fig.9 where we report the testing error on LFPW dataset (a subset of 300-W, see the beginning of this sec-tion) including 811 images for training and 224 for testing.

Note that knn-68 is in effect the full connection. Besides, we followed the observation in (Ren et al., 2014) and tried a shrinking settings ("knn-var" in the legend): knn-5 for the first third stages, knn-3 for the second third and knn-1 for the last third. This coarse-to-fine way is to capture the large displacement in early stages and to refine slightly in late stages. The specific settings in our experiments are $L = 68$, $M = 15$, $H = 5$, the number of stages $T = 24$.

As can be seen, smaller $k$ in knn tends to better results. However, the shrinking knn works best. This is consistent with the observations of (Ren et al., 2014) where the random pixel difference features are extracted with shrinking circle radius determined by cross-validation.

### 4.3. Results on Face Pose Data

We report the testing error on the face pose dataset 300-W (testing on common subset) as in Tab.1, where the parameters for our algorithm are $M = 15$, $T = 24$, shrinking knn as in Section 4.2. We also report the state-of-the-art results quoted from (Ren et al., 2014), where LBF and LBF fast were proposed in (Ren et al., 2014), SDM was proposed in (Xiong & De la Torre, 2013), . Our results seems inferior to the competitors. However, we conjecture that it's due to we are using a very simple regressor (i.e., a one hidden layer MLP) at each stage, while (Ren et al., 2014) utilized a tree ensemble with carefully tuned tree-depth and leaf values, which seems to have better generalization than a one hidden layer MLP. Also, we conjecture that a stronger PIF, e.g., one usees HoG as in (Xiong & De la Torre, 2013), would improve our results.

It would be our future work to test stage regressor $f(\cdot)$ with deeper structure. In the meanwhile, we anticipate that the proposed method in this paper would show its power with bigger training data as what has been observed for other deep neural network based methods, which would be also investigated in future work.

### 4.4. Results on 3D CT Images

As is discussed in the beginning of this section, the heart segmentation is converted to a 3D pose estimation problem, which goes similar with the 2D face pose estimation, except that for the testing error we adopt another measure that is more popular in medical image processing community (Zheng et al., 2008). Specifically, we calculate the so-called *mesh distance* between the predicted pose and the ground truth pose. Roughly speaking, the mesh distance is defined to indicate the closeness of two surfaces that are respectively comprised of triangles.

Following the convention of Medical Image literature, we report the mean mesh distance in millimeter over testing set for both the inner surface and the outer surface, as in Tab.2 where the baseline is "do nothing", i.e., the distance between initial guess and ground truth. We also quote a state-of-the-art result for heart segmentation based on AAM/ASM (Zheng et al., 2008). Note that the results there are obtained on their own proprietary datasets with a different scan parameters and image resolutions. The quoted results serve as a reference and should not be directly compared with ours.

## 5. Conclusions and Future Work

In this paper we propose a GTN representation for CPR so that it can be trained with BP globally which outperforms layer-wise (pre-)training. Sparse connection applied in order to learn local image feature representation. The proposed CPR-GTN is promising for 2D and 3D pose estimation. We'll explore the potential of CPR-GTN by testing it on much bigger data in the future.

# Globally Tuned Cascade Pose Regression via Back Propagation with Application in 2D face pose estimate and Heart Segmentation in 3D CT Volume: Supplement

**Peng Sun**
**James K Min**
**Guanglei Xiong**
Cornell

PES2021@MED.CORNELL.EDU
EMAIL@COAUTHORDOMAIN.EDU
EMAIL@COAUTHORDOMAIN.EDU

## 1. Implementation Details of Training

When extracting the random pixel difference features, the random points $z_1, ... z_i, ...$ are generated in a coordinate defined by mean face pose $\overline{p}$, where each point $z_i$ is anchored to its nearest landmark. For each training image $I$, we first determine the similarity transform $A$ from mean pose $\overline{p}$ to currently estimated pose $p$: $p = A(\overline{p})$, then we transform these points accordingly: $A(z_1), ..., A(z_i), ...$ which are the very points that we extract features for image $I$. This operation is more robust to rotation and scaling of face pose. See (Cao et al., 2014; Kazemi & Josephine, 2014) for details.

In line of (Dollár et al., 2010; Burgos-Artizzu et al., 2013; Cao et al., 2014; Kazemi & Josephine, 2014), we take a data augmentation trick which has been shown to improve CPR's performance.

We follow the suggestions in (Srivastava et al., 2014) for the NN regularization. Specifically, Rectified Linear Unit (Relu) is taken as activation function and Dropout with rate 0.5 is adopted. Stochastic Gradient Descent (SGD) is used with mini batch size 50. Parameters are updated with step size 0.1 and momentum 0.9. Other tuning parameters involving the CPR are discussed in the experiment section in the main paper.

## 2. Visualization: Synthesize Image from Pose

As a bonus of GTN representation, the trained CPR is rather easy to visualize (Simonyan et al., 2013), that is, for a given ground truth pose $p^*$, what's the most likely image $I$ the regressor $R(\cdot)$ is fed so that $p^* = R(\cdot)$? Details are put in supplement.

A trained CPR can be seen as a regressor mapping image to pose: $p = R(I)$. Now consider its inverse problem: given a ground truth pose $p^*$, what's the most likely image $I$ the regressor $R(\cdot)$ is fed? An interesting by-product of CPR-

GTN is that it makes such a visualization problem rather easy. Specifically, we want

$$\arg\min_I L = ||p^* - R(I)||_2^2. \tag{1}$$

Since we've formulated CPR as NN permitting BP, the above optimization problem can be solved by a technique introduced in (Simonyan et al., 2013). Basically, we can perform gradient descent with BP. To do this, we simply let the delta-signal pass all the way down to the variable $I$ and update $I$ with $-\eta \frac{\partial L}{\partial I}$ wehre $\eta$ is the step size. The initial value $I$ is set to the average face image over the training set.

# Updated Contents

## 1. Related Work

One of the ICML2015 reviewers pointed out that the idea of global tuning of CPR was already introduced in (Shi et al., 2014), which we were unaware of when we did this work during December 2014 to February 2015. The only difference with (Shi et al., 2014) is that we adopt Random Pixel Difference features and local connection, while (Shi et al., 2014) adopts HoG feature and fully connection. Besides, (Shi et al., 2014) tried to get rid of the initial pose guess by a global regression but saw a degraded performance.

## 2. Updated Experimental Results

After the submission of the preliminary manuscript to ICML2015, we continued our work and got some new results on 300-W datasets, as reported in Table 1. The improvement over our previous result is due to a more aggressive data augmentation (up to x200 expansion of the original training dataset, compared to x20 in our initial work) and a technique of summing up many mid-way losses as introduced in (Lee et al., 2014).

In Table 1 we can see that Deep Reg (Lee et al., 2014) is still better than this work, although they are based on almost the same framework. Our hunch for this difference is that HoG is used in Deep Reg, which could be more robust than Random Pixel Difference feature used in this work.

Note that every stage in CPR outputs a pose prediction. In Figure 1 we show the an example of these mid-way prediction with or without the technique in (Lee et al., 2014). We can see that the mid-way prediction is more interpretable after introducing the sum of the mid-way losses. Although the mid-way prediction does not directly relate to the final prediction at last stage, enforcing the mid-way result to be more like the ground truth during training seems a good regularization. See (Lee et al., 2014) for more details.

## 3. Code

Our initial code is based on Matlab CPU computation, which is very slow and prevents us from leveraging parameter tuning. We thus rewrite the code with GPU acceleration. The new code is available at `https://github.com/pengsun/bpcpr5`. The whole Directed Acyclic Graph (DAG) "framework" is built on top of a GPU accel-erated Matlab toolbox for Convolutional Network [1]. Additionally, we implement a GPU accelerated transformer for Random Pixel Difference feature extraction. The speed for our code is approximately 150 images per second with $T = 24$ stages on a typical contemporary GPU.

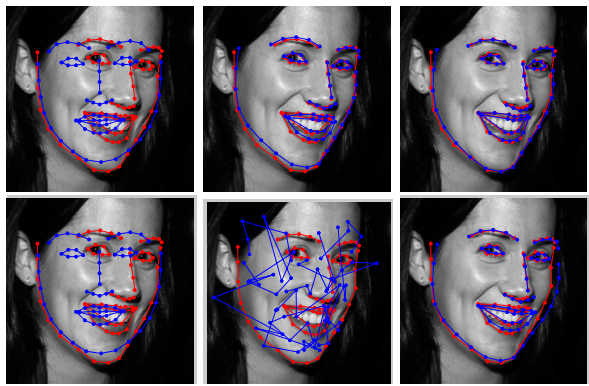

*Figure 1.* Intermediate results for tunable CPR. Red points: ground truth. Blue points: prediction. Column 1 to column 3: $T = 0$ (initial guess), $T = 4$ and $T = 24$. Top row: with suburgos2013m of mid-way losses. Bottom: only one loss transformer at the last stage.

*Table 1.* Average Distance on the testing set of 300-W (Normalized by Pupil Distance). The results for ESR, SDM, LBF (fast) are quoted from (Ren et al., 2014), Deep Reg from (Shi et al., 2014)

| Method | Fullset | Common-set | Challenging-set |
|---|---|---|---|
| ESR | 7.58 | 5.28 | 17.00 |
| SDM | 7.52 | 5.60 | 15.40 |
| LBF | 6.32 | 4.95 | 11.98 |
| LBF fast | 7.37 | 5.38 | 15.50 |
| Deep Reg | 6.31 | 4.51 | 13.80 |
| This work (new) | 7.46 | 5.24 | 16.56 |
| This work (old) | NA | 6.76 | NA |

---

[1] https://github.com/vlfeat/matconvnet